\pdfoutput=1

\documentclass[11pt]{article}

\usepackage{acl}

\usepackage{times}
\usepackage{latexsym}

\usepackage[T1]{fontenc}

\usepackage[utf8]{inputenc}

\usepackage{microtype}

\usepackage{times}
\usepackage{latexsym}
\usepackage{algorithm}
\usepackage{algorithmic}
\usepackage{amsmath}
\usepackage{amssymb}
\usepackage{graphicx}
\usepackage{color}
\usepackage{multicol}
\usepackage{multirow}
\usepackage{adjustbox}
\usepackage{subfigure}
\usepackage{hyperref}

\newcommand{\uw}{$^{\heartsuit}$}
\newcommand{\aitwo}{$^{\spadesuit}$}
\newcommand{\uu}{$^{\diamondsuit}$}

\title{Elaboration-Generating Commonsense Question Answering at Scale}

\author{Wenya Wang\uw \quad Vivek Srikumar\uu \aitwo \quad Hanna Hajishirzi\uw \aitwo \quad Noah A.~Smith\uw \aitwo \\
\uw{}Paul G. Allen School of Computer Science \& Engineering, University of Washington \\
\aitwo{}Allen Institute for AI \\
\uu{}School of Computing, University of Utah \\
  \texttt{wwenya@cs.washington.edu}
}
\begin{document}
\maketitle
\begin{abstract}

In question answering requiring common sense, language models (e.g., GPT-3) have been used to generate text expressing background knowledge that helps improve performance.  Yet the cost of working with such models is very high; in this work, we finetune smaller language models to generate useful intermediate context, referred to here as elaborations.  Our framework alternates between updating two language models---an elaboration generator and an answer predictor---allowing each to influence the other.  Using less than 0.5\% of the parameters of GPT-3, our model outperforms alternatives with similar sizes and closes the gap with GPT-3 on four commonsense question answering benchmarks. Human evaluations show that the quality of the generated elaborations is high.\footnote{The source code is available at \url{https://github.com/happywwy/Elabor/tree/main}.}


\end{abstract}

\section{Introduction}
\begin{figure}
\centering
\includegraphics[width=1.0\columnwidth]{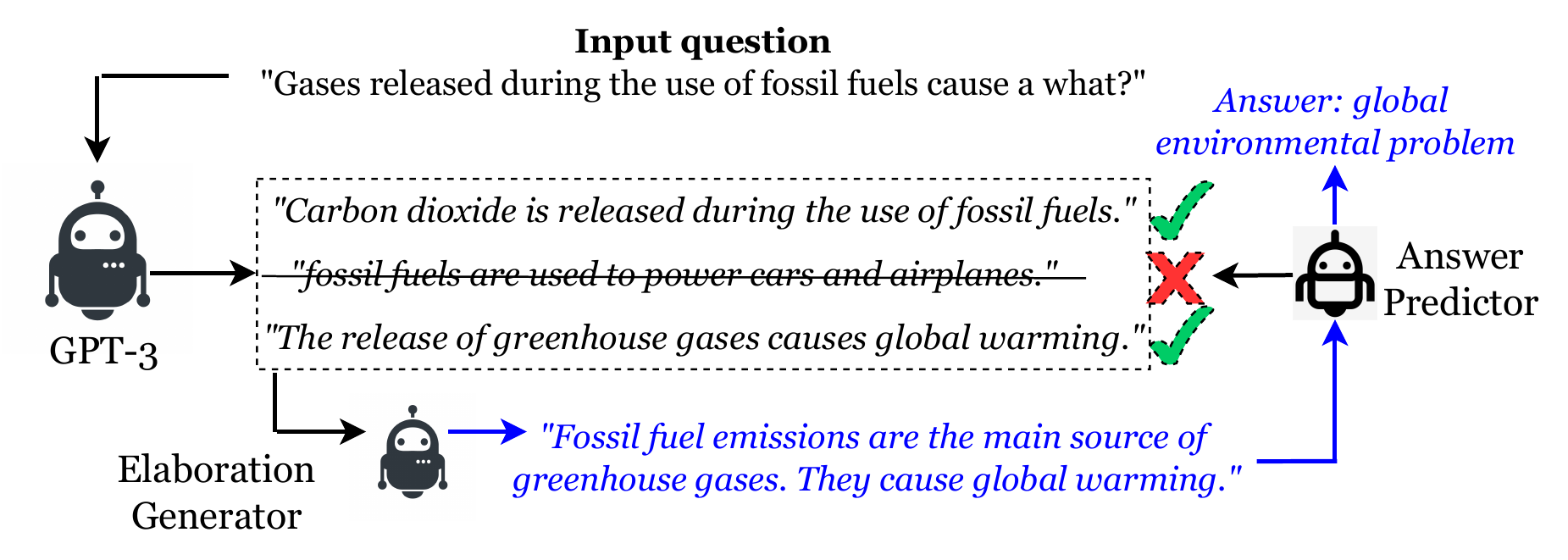}
\caption{An overview of the framework that selectively distills knowledge from GPT-3 to a smaller elaboration generator via an answer predictor.}\label{fig:teaser}
\vspace{-5mm}
\end{figure}

Commonsense question answering (QA; \citealp{talmor-etal-2019-commonsenseqa})  provides benchmarks used to evaluate the extent to which NLP models---increasingly based on language models---can ``understand'' questions and reason about their answers. For example, consider the question in Figure \ref{fig:teaser}: \textit{Gases released during the use of fossil fuels cause a what?} A reasonably informed human could give the answer \textit{global warming}, by reasoning that: \textit{Fossil fuel emissions are the main source of greenhouse gases. They cause global warming}. 

It is common to use LMs to predict answers directly for QA tasks \cite{devlin-etal-2019-bert,liu-et-al-roberta,khashabi-etal-2020-unifiedqa}. 
On challenging datasets whose questions rely on unstated background knowledge \cite{talmor2021commonsenseqa,mihaylov-etal-2018-suit,knot-et-al-qasc}, some recent works rely on external knowledge, e.g., Wikipedia or structured knowledge bases \cite{mihaylov-frank-2018-knowledgeable,lin-etal-2019-kagnet,banerjee-etal-2019-careful} for additional information that helps to answer the question. Such attempts are limited by the availability and coverage of the knowledge sources.
Another line of study \cite{liu-etal-2022-generated,paranjape-etal-2021-prompting,shwartz-etal-2020-unsupervised} reveals that generating text that expresses additional background knowledge relevant to a question is beneficial for answer prediction.
The ability to express such knowledge may promote model explainability by explicitly showing the reasoning process. However, expressing high-quality knowledge relies on  massive (and thus, expensive) pretrained LMs, e.g., GPT-3 with 175B parameters \cite{tom-et-al-gpt3}. 

In this work, we focus on a more practical setting and ask: Can smaller LMs, e.g., BART which is about 400$\times$ smaller than GPT-3, support reasoning and inference in an end-to-end manner?
To this end, we propose a scalable framework, alternating \textbf{ELAB}oration and answer predict\textbf{OR} (\textsc{ElabOr}), consisting of two interacting modules:  an elaboration generator and an answer predictor. Here an elaboration refers to additional context describing some background knowledge about the question. Instead of generating elaborations independently, we propose a probabilistic framework that treats the elaboration as a latent variable and iteratively optimizes the elaboration generator after receiving feedback from the answer prediction.  
Specifically, for each question-answer pair $(q,a)$, we decompose the distribution of the answer conditioned on the question $P(a\mid q)$ into a distribution $P(e\mid q)$ over a latent elaboration, modeled by the \textbf{elaboration generator}, and a likelihood distribution $P(a \mid e,q)$ over the answer, modeled by the \textbf{answer predictor}. We alternately train the elaboration generator and the answer predictor so that each can benefit the other.  Earlier work either pre-constructs elaborations $e$  from external knowledge~\cite{mihaylov-frank-2018-knowledgeable}  or learns $P(e\mid q)$  solely based on annotations \cite{rajani-etal-2019-explain}; we learn the elaboration generator by distilling high-quality knowledge from GPT-3. We do this using a procedure inspired by hard Expectation-Maximization \cite{MinCHZ19}. This involves refining and filtering elaborations informed by the answer predictor, as shown in Figure~\ref{fig:teaser}.  \textsc{ElabOr} is thus capable of propagating information in both directions: from elaboration generator to answer predictor and vice versa.

We conduct experiments on four commonsense QA datasets: CommonsenseQA \cite{talmor-etal-2019-commonsenseqa}, CommonsenseQA 2.0 \cite{talmor2021commonsenseqa}, Scientific Commonsense \cite{knot-et-al-qasc}, and OpenBookQA \cite{mihaylov-etal-2018-suit}. Our experiments reveal that (1) alternating training with smaller LMs (e.g., BART, and GPT-2) narrows the gap between small models and GPT-3; (2) the ability to generate and reason with background elaborations indeed brings larger performance gains than direct inference on more challenging Commonsense QA datasets; (3) the alternating framework helps to filter irrelevant elaborations generated from GPT-3 and the learned elaboration generator can express information that helps to answer the question, as shown through human evaluations.

\section{Modeling Answers and Elaborations}\label{sec:formulation}

We focus on the task of commonsense question answering in the multiple-choice setting: we seek to identify the answer to a commonsense question among provided candidate choices. Importantly, we are not provided with additional elaboration that may be needed to do so.
We formalize the setting and define the model in this section, and Section \ref{sec:alternating} details the training procedure.

\subsection{Elaborations as a Latent Variable}

We formalize commonsense QA in a probabilistic framework. Given a question $q$ and its correct answer $a$, we seek to train a model that maximizes the probability of the correct answer $P(a\mid q)$. Directly predicting the answer can be be challenging when complex understanding is needed. Moreover, doing so renders the provenance of the answer unclear. 
To address both issues, we assume that the answer depends on some latent elaboration $e\in E$ with $E$ denoting a set of probable elaborations.
With the latent variable, the training objective becomes 
\begin{equation}\label{eqn:overall}
    \log P(a\mid q) = \log \sum_{e\in E} P(e \mid q) P(a\mid e, q).
\end{equation}
Here, the first term in the summation, $P(e\mid q)$, denotes the probability of an elaboration $e$ conditioned on question $q$ and is captured by the \emph{elaboration generator}. The second term $P(a\mid e, q)$ characterizes the distribution of the answer $a$ conditioned on both the elaboration and the question and is captured by the \emph{answer predictor}. The decomposition in Eq. \ref{eqn:overall} has also been adopted by \citet{Lewis2020}, taking retrieved knowledge as the hidden variable. Different from the retrieval setting, the generation distribution $P(e\mid q)$ is intractable. We instead resort to hard EM and alternating optimization.

\subsection{A Joint Model}\label{sec:joint_model}
The elaboration generator seeks to generate an elaboration sequence $e$ given the question $q$ as a prompt. 
We denote the conditional probability of an elaboration given a question by $\mathcal{F}_E$; that is, using the notation from Eq.~\ref{eqn:overall}, we have $P(e\mid q) = \mathcal{F}_E(e, q; \Phi)$. 
We model the elaboration generator using a generative language model that computes the distribution of tokens at each generation step: 
\begin{equation}
    \mathcal{F}_E(e, q; \Phi) = \prod_{t=1}^{m} p_{\mathtt{GEN}}(e_t\mid q,e_1,...,e_{t-1}),
\end{equation}
where $e=\{e_1,...,e_m\}$ denotes the generated elaboration sequence. In our experiment, we adopt two generation models---BART \cite{lewis-etal-2020-bart} and GPT-2 \cite{Radford2019}---to model $p_{\mathtt{GEN}}$.

The answer predictor, denoted  $\mathcal{F}_{A}$, aims to produce the probability of an answer sequence $a$ given a question $q$ and an elaboration $e$, i.e., $P(a\mid e,q) = \mathcal{F}_{A}(a, e, q; \Theta)$. Any language model could be adopted as the answer predictor. For generality, we select two commonly-used language models from two different paradigms, namely BERT \cite{devlin-etal-2019-bert} as a masked language model and T5 \cite{raffel-et-al-t5} as a generative language model. For T5, $\mathcal{F}_{A}(a, e, q; \Theta)$ is computed for an answer sequence $a=\{a_1,...,a_n\}$ using
\begin{equation}
    \mathcal{F}_{A}(a, e, q; \Theta) = \prod_{t=1}^{n} p_{\mathtt{T5}}(a_t\mid e,q,a_1,...,a_{t-1}),
\end{equation}
with $p_{\mathtt{T5}}$ denoting the generation probability of token $a_t$ using T5. For BERT, $\mathcal{F}_{A}(a, e, q; \Theta)$ is computed using a softmaxed linear layer over the representation of the [CLS] token:
\begin{eqnarray}
    \mathcal{F}_{A}(a, e, q; \Theta) = \mathrm{softmax}(\mathbf{W} \mathbf{h}_{[CLS]} + \mathbf{b})
\end{eqnarray}
by giving ``[CLS] elaboration [SEP] question [SEP] answer [SEP]'' to BERT.

\subsection{Inference}\label{sec:inference}
In the testing phase, for each question, we first use the trained elaboration generator $\mathcal{F}_E$ to sample a set of elaborations $\tilde{\mathcal{E}}$. For each $\tilde{e}\in\tilde{\mathcal{E}}$, we use the answer predictor $\mathcal{F}_A$ with softmax to produce a normalized distribution over the candidate set. 
By running the answer predictor for each sampled elaboration, we take the maximum probability as the score for candidate $a^i$ which is then used to produce the final prediction:
\begin{equation}\label{eqn:inference}
    a' = \mathrm{argmax}_{a^i\in \mathcal{A}} \max_{\tilde{e}\in\tilde{\mathcal{E}}} \frac{\exp^{\mathcal{F}_A(a^i, \tilde{e}, q; \Theta)}}{\sum_{a^j\in \mathcal{A}} \exp^{\mathcal{F}_A(a^j, \tilde{e}, q; \Theta)}}
\end{equation}
with $\mathcal{A}$ denoting the set of candidate answers.

\section{Alternating Elaboration and Answer Predictor (\textsc{ElabOr})}\label{sec:alternating}

Many existing retrieval or knowledge-based QA methods only optimize $P(a \mid e, q)$, assuming $e$ is given and fixed. Explanation-based methods, on the other hand, train $P(e \mid q)$ separately using human-annotated explanations. Doing so poses two problems: (1) we need an annotated explanation corpus, and (2) the elaboration generator cannot be calibrated towards the answer. 

\begin{figure}
\centering
\includegraphics[width=1.0\columnwidth]{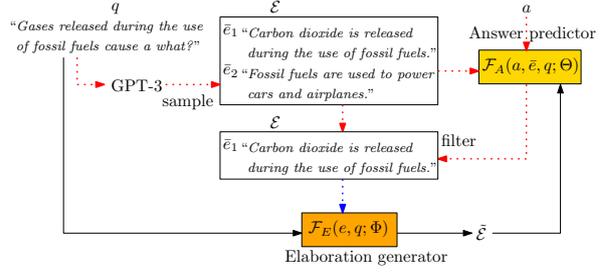}
\caption{The training framework, which alternates between learning the elaboration generator (dotted arrows) and learning the answer predictor (solid arrows). The elaboration generator is optimized via an EM-like algorithm with the E-step (\textcolor{red}{red arrow}) sampling and filtering high-quality elaborations and the M-step (\textcolor{blue}{blue arrow}) maximizing the probability of $\mathcal{E}$.}\label{fig:overall}
\end{figure}

In this work, we propose an approach that tackles both problems by jointly training the elaboration generator and the answer predictor in an alternating framework. Figure~\ref{fig:overall} illustrates the overall architecture for training. In each iteration, the elaboration generator $\mathcal{F}_E$ learns to produce high-quality elaborations using feedback from the answer predictor (Section \ref{sec:em}). The answer predictor $\mathcal{F}_A$ then takes the generated elaborations as input to produce more reliable answers (Section \ref{sec:answer}). This strategy allows  mutual interaction between the two components, propagating information in both directions.  To reduce the search space of possible elaborations, we propose to distill knowledge from the pretrained GPT-3 model in a selective way to learn a lightweight elaboration generator (Section~\ref{sec:distill}).

\subsection{An EM-Inspired Learner}\label{sec:em}
Our goal is to optimize  Eq.~\ref{eqn:overall}, rewritten below:
\begin{equation}\label{eqn:expectation}
    \log P(a\mid q) = \log \mathbb{E}_{e\sim P(e\mid q)} [P(a\mid e,q)].
\end{equation}
Directly optimizing the elaboration generator in this expression is difficult.\footnote{One popular  option would be to adopt the REINFORCE algorithm \cite{Williams:92} that updates $\mathcal{F}_E(e, q; \Phi)$ using differentiable policy gradient. However, this strategy involves searching in a huge symbolic space and can be unstable.} Inspired by \citet{qu2021rnnlogic}, we adopt a hard EM framework to do so. The E-step first generates a set of elaborations related to the question and then selects ``good'' elaborations that help to predict the correct answer. The M-step maximizes the probability of generating these ``good'' elaborations. 

\vspace{2mm}
\noindent\textbf{E-Step.} The E-step aims to identify a set of ``good'' elaborations from the posterior probability of an elaboration $e$ after observing the correct answer $a$:
\begin{equation}\label{eqn:posterior}
    P(e\mid q,a) \propto P(e\mid q) P(a\mid e,q)
\end{equation}
The posterior approximation on the right-hand-side of Eq. \ref{eqn:posterior} aligns with the intuition that the elaboration could have higher probability if it is both relevant to the question (i.e., $P(e\mid q)$) and, when combined with the question, provides higher chance of predicting the correct answer (i.e., $P(a\mid e,q)$). 

However, the intractable space of possible elaborations renders sampling from  $P(e \mid q) P(a\mid e,q)$ nontrivial.
To alleviate this issue, we adopt two approximations. First, we use GPT-3 to produce more reliable distribution $P(e \mid q)$, and thus rewriting Eq.~\ref{eqn:posterior} as $P(e\mid q,a) \propto P_{\texttt{GPT-3}}(e \mid q) P(a\mid e,q)$. Second, we approximate the sampling process via a two-step sample-and-filter procedure. Specifically, we first sample a set of elaborations $\bar{\mathcal{E}}$ from $P_{\texttt{GPT-3}}(e\mid q)$ which will be discussed in Section \ref{sec:distill}. Then, we filter $\bar{\mathcal{E}}$ according to $P(a\mid e,q)$. Specifically, for each $\bar{e}\in\bar{\mathcal{E}}$, we use the answer predictor\footnote{We also study other filtering strategies as detailed in Section \ref{sec:analysis}.} to produce $P(a\mid \bar{e},q) = \mathcal{F}_A(a, \bar{e}, q)$. Then we select top-$K$ elaborations from $\bar{\mathcal{E}}$ to form $\mathcal{E}$ as the set of ``good'' elaborations. This operation allows the answer predictor to assist in learning how to select elaborations.

\vspace{2mm}
\noindent\textbf{M-Step.} With the selected context set $\mathcal{E}$ produced in the E-step, the M-step aims to maximize the probability of each elaboration $e\in\mathcal{E}$ to update the elaboration generator $\mathcal{F}_E$ while keeping the answer predictor fixed:
\begin{equation}\label{eqn:maxFc}
    \max_{\Phi} \log P(\mathcal{E}\mid q) = \max_{\Phi} \sum_{e\in\mathcal{E}} \log \mathcal{F}_E(e,q;\Phi),
\end{equation}
given $P(\mathcal{E}\mid q)=\prod_{e\in\mathcal{E}} P(e|q)$. In this way, the elaboration generator learns to produce elaborations that are both relevant to the question and with a higher probability of predicting the correct answer. Eq.~\ref{eqn:maxFc} could also be viewed as a kind of selective distillation, which instead of distilling all the sampled elaborations $\bar{\mathcal{E}}$ from GPT-3, learns to filter out noisy elaborations before transferring knowledge to the elaboration generator.

\vspace{2mm}
\subsection{Optimizing Answer Predictor}\label{sec:answer}
After updating the elaboration generator, the next step of the alternative training aims to update the answer predictor $\mathcal{F}_{A}(a, e, q; \Theta)$ while keeping the elaboration generator fixed. To achieve that, we approximate the objective of Eq.~\ref{eqn:expectation} to $\log P(a \mid \tilde{e}, q)$ by sampling a set of elaborations $\tilde{e}\in\tilde{\mathcal{E}}$ from the elaboration generator $P(\tilde{e}\mid q)=\mathcal{F}_E(\tilde{e}, q; \Phi)$. 
Then the objective becomes to maximize
\begin{equation}\label{eqn:predictor}
    \log P(a\mid \tilde{e}, q) = \log \mathcal{F}_{A}(a, \tilde{e}, q; \Theta)
\end{equation}
for the correct answer $a$. The sampled elaboration $\tilde{e}$ from the elaboration generator acts as additional background and explanation for the question, which helps to learn a more reliable prediction model to answer the question. 
The alternation between updating the answer predictor and the elaboration generator promotes mutual enhancement of each component.
The entire training procedure of \textsc{ElabOr} can be found in Appendix \ref{sec:app-algo}.

\subsection{Distilling GPT-3}\label{sec:distill}
As discussed in the E-step, we use GPT-3\footnote{We also tried more accessible models, e.g., GPT-J (6B), but observed much worse generation quality.} to sample possible elaborations to train our elaboration generator.
\citet{liu-etal-2022-generated} showed that, using a small number of prompts and a question, GPT-3 can generate useful knowledge to enhance answer prediction. Inspired by \citet{hinton2015distilling} and \citet{West2021SymbolicKD}, we adopt the idea of knowledge distillation to transfer knowledge from GPT-3 (expensive to deploy at inference time) to our (cheaper) elaboration generator. We first use GPT-3 to generate a set of elaborations given some predefined prompts. Following \citet{liu-etal-2022-generated}, for each task, we design the prompt as a short instruction followed by five demonstrative examples and a new-question placeholder. By plugging each question into the placeholder, we can repeatedly sample an elaboration $\bar{e}$ as the continuation of the prompt.  This yields a set of candidate elaborations, $\bar{\mathcal{E}}$. 

Here we use nucleus sampling \cite{Holtzman2020The} to sample each elaboration $\bar{e}$. For knowledge distillation, a naive strategy could be optimizing the elaboration generator by minimizing
\begin{equation}
    D(P_{\texttt{GPT-3}}, P_{s}) = \mathbb{E}_{\bar{e}\sim P_{\texttt{GPT-3}}} [-\log P_{s}(\bar{e} \mid q)], \nonumber
\end{equation}
with $P_s$ denoting the student network, i.e., our elaboration generator. However, as shown in the experiments, GPT-3 is prone to generating noisy text sequences that may not be relevant to answer the question. This would lead to negative transfer. Our proposal in the E-step is a form of selective knowledge distillation \cite{kang-etal-2020-regularization} which filters elaborations generated from GPT-3 according to the answer score before optimizing our student model.

\section{Experiments}

In this section, we examine the question: \emph{Does jointly optimizing the elaboration generator with the answer predictor outperform approaches that merely retrieve knowledge from trained models, if at all?} As a secondary objective, we also investigate the impact of the design choices in our approach, including the choice of the language model, the need for distillation, the choice of elaboration filtering and the decoding strategy. 

\subsection{Data and Setup}
We select four multiple-choice commonsense QA datasets involving commonsense concepts or scientific facts: (1) CommonsenseQA (\textbf{CSQA}; \citealp{talmor-etal-2019-commonsenseqa}), (2) CommonsenseQA 2.0 (\textbf{CSQA2},\citealp{talmor2021commonsenseqa}) (3) Scientific Commonsense (\textbf{QASC}, \citealp{knot-et-al-qasc}), and (4) OpenBookQA (\textbf{OBQA}; \citealp{mihaylov-etal-2018-suit}). The elaboration generator is implemented using GPT2-large \cite{Radford2019} and BART-large \cite{lewis-etal-2020-bart}. The answer predictor is implemented using T5-large \cite{raffel-et-al-t5} and BERT-base-uncased \cite{devlin-etal-2019-bert}. We also experiment with more competitive and larger  answer predictors, e.g., UnifiedQA-large/3b \cite{khashabi-etal-2020-unifiedqa}. We sample 20 elaborations from GPT-3, of which 3 are selected to form $\mathcal{E}$. We sample 10 elaborations from our elaboration generator during both training and inference. Appendix \ref{sec:setup} has more details on the datasets and experiment settings.

\begin{table}
\centering
	\begin{adjustbox}{max width=1.0\columnwidth}
\begin{tabular}{l|c|cc|cc|cc}
\hline
\textbf{Dataset}  & \textbf{CSQA}  & \multicolumn{2}{c|}{\textbf{CSQA2}}  & \multicolumn{2}{c|}{\textbf{QASC}}  & \multicolumn{2}{c}{\textbf{OBQA}}  \\ 
$\mathcal{F}_A$ & T5-large & \multicolumn{2}{c|}{T5-large} & \multicolumn{2}{c|}{T5-large} & \multicolumn{2}{c}{BERT}  \\ 
Eval set & dev. & dev. & test & dev. & test & dev. & test \\ \hline
 vanilla             & 65.19    & 55.25  & 54.91   & 48.49 & 45.22   & 54.80 & 51.00 \\ \hline  
COMET      & 66.34    & 52.11  & -   & 49.35  & -  & 55.00 & - \\ 
Wikipedia  &   63.14       &     52.14  & -   &     48.16  & -   &   54.20  & -  \\ \hline
selftalk   & 65.03    & 55.88  & 54.87  & 50.22  & 46.85  & 53.60 & 54.40 \\ 
GPT-3       & 67.23    & 58.56  & 56.98   & \textbf{55.18}  & \textbf{53.04}   & \textbf{58.60} & \textbf{59.40} \\ \hline
\multicolumn{8}{c}{Elaboration model: GPT2-large} \\ \hline 
scratch & 65.36    & 56.99  & -   & 50.65  & -  & 55.80 & - \\ 
pipeline     & 66.42    & 56.63  & 53.54  & 52.48  & 49.13  & 56.60 & 55.00 \\  
\textsc{ElabOr}         & \textbf{67.32}    & \textbf{58.72}  & \textbf{57.58}  & 54.21  & 50.22  & \textbf{58.60} & 56.40 \\ \hline
\end{tabular}
\end{adjustbox}
\caption{Accuracies for the proposed model and baselines. GPT2-large is used as the elaboration generator.}\label{tbl:baseline}
\end{table}

\begin{table}
\centering
\begin{adjustbox}{max width=1.0\columnwidth}
\begin{tabular}{l|cc|cc|cc|cc}
\hline
\textbf{Dataset}  & \multicolumn{2}{c|}{\textbf{CSQA}}  & \multicolumn{2}{c|}{\textbf{CSQA2}} & \multicolumn{2}{c|}{\textbf{QASC}}  & \multicolumn{2}{c}{\textbf{OBQA}}  \\ 
Generator & BART & GPT2 & BART & GPT2 & BART & GPT2 & BART & GPT2  \\ \hline
scratch  & 64.29    & 65.36    & 55.45 & 56.99   & 49.14 & 50.65 & 55.80 & 55.80 \\  
pipeline      & 65.60 & 66.42    & 56.47 & 56.63    & 51.73 & 52.48    & 56.40 & 56.60 \\ 
\textsc{ElabOr}   & 66.26  & \textbf{67.32}  & 58.09 & \textbf{58.72}    & 53.78 & \textbf{54.21}   & 57.60 & \textbf{58.60} \\ \hline
\end{tabular}
\end{adjustbox}
\vspace{-2mm}
\caption{Results on dev.~set for different context generators: BART-large and GPT2-large.}\label{tbl:generator}
\vspace{-4mm}
\end{table}

\subsection{Baselines}
We organize the baselines into four groups: (1) Direct answer prediction without additional knowledge (\textbf{vanilla}). (2) Answer prediction with retrieved knowledge: \textbf{COMET} \cite{bosselut-etal-2019-comet} is trained on the ATOMIC corpus \cite{sap-et-al-atomic} to automatically generate causes and effects of a question. 
\textbf{Wikipedia} follows \citet{chen-etal-2017-reading}, which retrieves and ranks text spans in Wikipedia articles. 
(3) Fixed elaboration generator: \textbf{selftalk} generates extra background knowledge based on some clarification questions \cite{shwartz-etal-2020-unsupervised}. 
\textbf{GPT-3} \cite{tom-et-al-gpt3} samples 10 knowledge spans as continuations of the question using some demonstrative prompts. (4) Trained elaboration generator: \textbf{scratch} implements alternative training without distilling knowledge from GPT-3. \textbf{pipeline} first pretrains the generator using all the sequences generated from GPT-3, then finetunes the answer predictor. 
For fair comparisons, all four groups require training the answer predictor $\mathcal{F}_A$. The second and third groups additionally involve intermediate contexts which are kept fixed. The last group learns both an elaboration generator and an answer predictor. During inference, we pick the choice with maximum score across all the knowledge sequences or generations following Eq.~\ref{eqn:inference}.

\begin{table}
\centering
\begin{adjustbox}{max width=1.0\columnwidth}
\begin{tabular}{l|cc|cc|cc|cc}
\hline
\textbf{Dataset}  & \multicolumn{2}{c|}{\textbf{CSQA}}  & \multicolumn{2}{c|}{\textbf{CSQA2}} & \multicolumn{2}{c|}{\textbf{QASC}}  & \multicolumn{2}{c}{\textbf{OBQA}}  \\ 
Predictor & T5-id & U-3b & T5-id & U-3b & T5-id & U-3b & T5-id & U-3b  \\ \hline
vanilla  & 70.43    & 81.41    & 54.94 & 64.46   & 57.56 & 74.73 & 68.20 & 79.60 \\  
GPT-3      & 75.68 & \textbf{81.90}    & 55.73 & \textbf{67.30}    & 64.69 & \textbf{77.11}    & 74.40 & 82.40 \\ 
GenMC  & 72.67 & - & - & - & 58.06 & - & 71.60 & -\\
\textsc{ElabOr}   & 74.61  & 81.10  &  57.62  &  65.53   & 64.04 & 76.78   & 73.20 & \textbf{83.80} \\ \hline
\end{tabular}
\end{adjustbox}
\caption{Results for T5-large with answer IDs as outputs (T5-id) and UnifiedQA-3b (U-3b) as answer predictors.}\label{tbl:predictor}
\vspace{-2mm}
\end{table}

\subsection{Results}
Table~\ref{tbl:baseline} shows the main experimental results. Here we use T5-large as the answer predictor for CSQA, CSQA2, QASC, and BERT for OBQA. These are chosen according to the best performances given. To account for more general scenarios, we first use T5 in an open-domain QA setting where no answer choices are given as input, and the target output is the gold answer tokens. We also experiment with other input/output formats for T5 as will be shown in Section \ref{sec:analysis}.
From Table~\ref{tbl:baseline}, the advantage of additional knowledge or elaborations is more evident for CSQA2, QASC, and OBQA, compared with CSQA (which contains relatively simpler questions). This confirms the importance of reasoning for complex QA problems.
GPT-3 demonstrates performance gains over other knowledge sources. Using less than 5\% of the parameters of GPT-3, \textsc{ElabOr} outperforms GPT-3 on two datasets. 
It also clearly outperforms those models having similar computational cost (e.g., scratch, pipeline). The performance gain of \textsc{ElabOr} over pipeline demonstrates the advantage of our alternating framework. The scratch model on the other hand is prone to learning meaningless shortcuts, e.g., ``\textit{The correct answer: I know I'm not sure but whatever}.''

\subsection{Analysis}\label{sec:analysis}
In subsequent experiments, we use the development set of each corpus to make evaluations because the test set is not publicly available.

\noindent\textbf{Elaboration Generator.} Table \ref{tbl:generator} shows the effects of different LMs, specifically BART-large and GPT2-large, as elaboration generators. Both demonstrate consistent results across different training strategies (scratch, pipeline, \textsc{ElabOr}). In addition, GPT2-large slightly outperforms BART-large across all the experiments. The higher performance of GPT2-large could be credited to a larger parameter size (774M) compared to BART-large (406M). Another observation is that GPT2-large has more generation flexibility which appears to be less repetitive and cover more aspects relevant to the question, compared to BART-large. 

\noindent\textbf{Answer Predictor.} Table \ref{tbl:predictor} reveals the effect of our framework on more competitive settings and larger answer predictors. We consider another input/output format for T5, referred to as T5-id, which takes both IDs (we use (A), (B), etc. as answer IDs) and tokens of the answer choices as input, and the ID for the gold answer as output. This was adopted in GenMC \cite{huang-etal-2022-clues}. Obviously, T5-id outperforms T5 under the open-domain setting (Table \ref{tbl:baseline}) by a large margin, and \textsc{ElabOr} shows clear gains over GenMC. A larger model, UnifiedQA-3b, brings huge improvements even for the vanilla model. Still, additional elaborations (GPT-3 or \textsc{ElabOr}) bring further improvements across all the datasets.

\begin{table}
\centering
	\begin{adjustbox}{max width=0.9\columnwidth}
\begin{tabular}{l|l|c|c|c|c}
\hline
\textbf{Setting}  & \textbf{Variants}  & \textbf{CSQA}  & \textbf{CSQA2}  & \textbf{QASC}  & \textbf{OBQA}  \\ \hline
\multirow{4}{*}{\begin{tabular}[l]{@{}l@{}}\textbf{Elaboration}\\\textbf{filtering}\end{tabular}}   & random    & 66.34  &  57.58  &  52.27   & 55.40 \\ 
& correct   & 66.34  &  57.97  & 54.10  & 56.20  \\ 
& pos-neg  &  66.58   &  \textbf{58.72}   &  54.00    &  58.20   \\ 
& \textbf{pos}      &  \textbf{67.32}  &  \textbf{58.72}  &  \textbf{54.21}  &  \textbf{58.60} \\ \hline
\multirow{4}{*}{\begin{tabular}[l]{@{}l@{}}\textbf{Elaboration}\\\textbf{integration}\end{tabular}}  & concatenate & 50.86  & 55.92  &  40.39  & 57.20 \\ 
& probability   &  65.19  &  57.58  &  52.48  & 57.60 \\  
& similarity  &  65.77  &  56.47  &  52.16  &  \textbf{59.40} \\ 
& \textbf{maximum}  &  \textbf{67.32}  &  \textbf{58.72}  &  \textbf{54.21}  &  58.60 \\ \hline
\multirow{3}{*}{\begin{tabular}[l]{@{}l@{}}\textbf{Elaboration}\\\textbf{generation}\end{tabular}}  & greedy & 64.13  & 55.14  &  50.86  & \textbf{59.00} \\ 
& beam   &  66.01  & 57.97  & 52.70  & 58.80 \\  
& \textbf{sample}  & \textbf{67.32}  &  \textbf{58.72}  &  \textbf{54.21}  &  58.60   \\ \hline
\end{tabular}
\end{adjustbox}
\vspace{-2mm}
\caption{Results of model variations: (1) changing elaboration filtering criteria during E-step; (2) changing elaboration integration methods for inference; (3) changing generation settings for GPT2-large.}\label{tbl:setting}
\vspace{-4mm}
\end{table}

\vspace{1mm}
\noindent\textbf{Elaboration Filtering}. The first block (Elaboration filtering) of Table \ref{tbl:setting} shows the effect of different filtering criteria as discussed in the E-step of Section \ref{sec:em}. We implement three other filtering strategies. The \textbf{random} option filters GPT3-generated elaborations by randomly selecting 3 out of 20. The \textbf{correct} option selects all the elaborations that produce the correct answer when fed into the answer predictor. The \textbf{pos-neg} option computes the score difference between the correct answer and the average of incorrect answers, based on which 3 elaborations with highest scores are being selected. The \textbf{pos} option uses the answer predictor as adopted by \textsc{ElabOr}. Clearly, random selection produces inferior results among all the options, verifying the benefit of filtering high-quality elaborations for training the elaboration generator. 

\vspace{1mm}
\noindent\textbf{Elaboration Integration}. The second block (Elaboration integration) of Table \ref{tbl:setting} investigates the effect of different elaboration integration methods during inference. Recall from Eq.~\ref{eqn:inference} that \textsc{ElabOr} uses \textbf{maximum} pooling among all the generated elaborations $\tilde{\mathcal{E}}$ for final predictions. We are interested in how different inference strategies may affect the final performance. Specifically, instead of maximum pooling, we concatenate all the elaborations in $\tilde{\mathcal{E}}$ in a single sequence and feed it into the answer predictor (\textbf{concatenate}). This brings a clear performance drop on CSQA and QASC, probably due to the unexpected noise and the forgetting issue for long sequences. Another strategy is to formalize inference with a probabilistic view where each generated elaboration has a probability contributing to the final prediction via weighted aggregation (\textbf{probability}). To produce the probability, we apply a softmax layer on top of the output logit of each generated elaboration $\tilde{e}\in\tilde{\mathcal{E}}$. The last option is to compute the similarity between each elaboration and the question and use the most similar elaboration for final inference (\textbf{similarity}). We use sentence embeddings generated from sentence transformers \cite{reimers-gurevych-2019-sentence} with cosine similarity to select the optimal elaboration. As a result, maximum pooling outperforms other variations at most of the times.

\vspace{1mm}
\noindent\textbf{Decoding Strategy}. The last block (Elaboration generation) of Table \ref{tbl:setting} reflects how different decoding strategies inherent in the LMs may affect the final performance. We compare the results of greedy decoding (\textbf{greedy}) where each decoding step only selects the token with highest probability, beam search (\textbf{beam}) with size 10 at each decoding step and selecting top 10 sequences via nucleus sampling (\textbf{sample}) adopted in the proposed model \textsc{ElabOr}. Clearly, decoding via sampling produces the best results or comes very close.

\begin{figure}
\centering
\includegraphics[width=1.0\columnwidth]{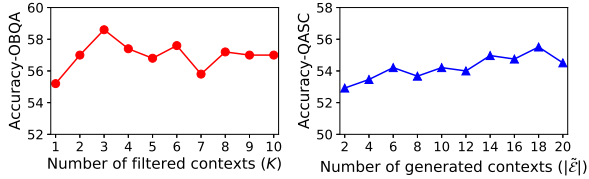}
\vspace{-5mm}
\caption{Sensitivity analysis of \textsc{ElabOr}. The left figure depicts results on OBQA when varying the number of selected elaborations from GPT-3. The right figure depicts results on QASC when varying the number of generated elaborations.}\label{fig:sens}
\vspace{-4mm}
\end{figure}

\vspace{1mm}
\noindent\textbf{Sensitivity Test}. Figure \ref{fig:sens} demonstrates the effects of changing (1) the number of filtered high-quality elaborations ($K$) from GPT-3 and (2) the size of set $\tilde{\mathcal{E}}$ corresponding to the total number of elaborations generated from the elaboration generator. The left plot demonstrates the performance increases when increasing $K$ from 1 to 3, but decreases for $K>3$. This pattern verifies that GPT-3 may generate elaborations that negatively affect the final performance. On the other hand, increasing the number of sampled elaborations from the elaboration generator (from 2 to 20) during both training and testing phases brings gradual improvements. This is as expected, given that sampling a diverse set of elaborations should add up to a wide coverage of relevant knowledge for the question. 

\subsection{Human Evaluation}
To evaluate the quality of elaborations for question answering, we conduct two sets of human evaluations on QASC and CSQA2. For the first experiment, we investigate whether the filtered elaborations from GPT-3 are considered more helpful to answer the question compared to those that are not selected by the model. For the second experiment, we evaluate the quality of the generated elaborations. Some concrete examples of questions and generations can be found in Appendix \ref{sec:app-generations}. The annotation task was carried out in Amazon Mechanical Turk. We restrict annotators to those located in English-speaking countries and who have at least 99\% approval rate over more than 1000 tasks. The results are aggregated using majority vote among annotations from 3 workers. Our institution's IRB approved the study.   We paid workers an estimated US\$15 per hour. 

\begin{figure}
\centering
\includegraphics[width=0.9\columnwidth]{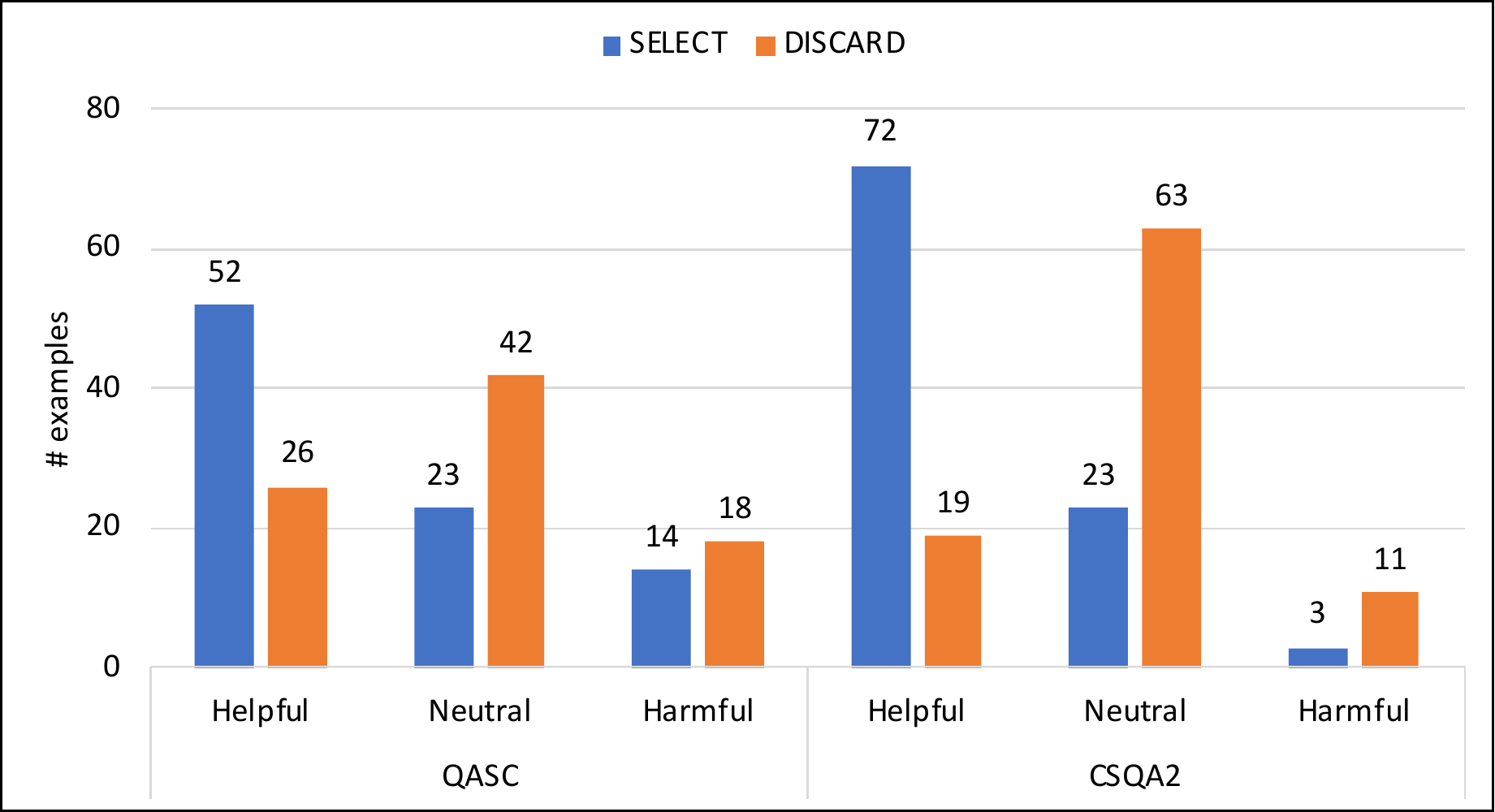}
\caption{Human evaluation results for SELECT and DISCARD elaborations generated by GPT-3.}\label{fig:human1}
\vspace{-2mm}
\end{figure}

\vspace{1mm}
\noindent\textbf{Effect of Filtering.} Recall that we use the answer predictor to filter elaborations generated from GPT-3 in the E-step. To demonstrate whether the filtering process is capable of removing noisy elaborations, we randomly sample 100 questions from the training corpus of each of two datasets (QASC, CSQA2). For each instance, we present the crowd workers with a question, the correct answer, the GPT3-generated elaboration $e$ that has the highest score $P(a\mid e,q)$ (denoted  SELECT), and an elaboration randomly sampled from the remaining ones that are discarded by the answer predictor (denoted DISCARD). The workers are then asked to evaluate the SELECT and DISCARD elaborations by choosing 1-out-of-3 choices: \emph{helpful} (the elaboration adds useful information to answer the question), \emph{neutral} (the elaboration has no influence on the problem), and \emph{harmful} (the elaboration is misleading). To avoid annotation bias, we randomize the order of SELECT and DISCARD elaborations for each example. The results are shown in Figure \ref{fig:human1}. Among 100 examples for each dataset, the number of helpful elaborations annotated by the workers is considerably higher for the selected category than that of the discarded category. In contrast, the workers agree that the selected elaborations are less likely to be neutral or harmful compared to those that are discarded. The difference is even more evident on CSQA2. This verifies the necessity of using the answer predictor to filter noisy elaborations generated by GPT-3 before distilling the knowledge.

\begin{figure}
\centering
\includegraphics[width=1.0\columnwidth]{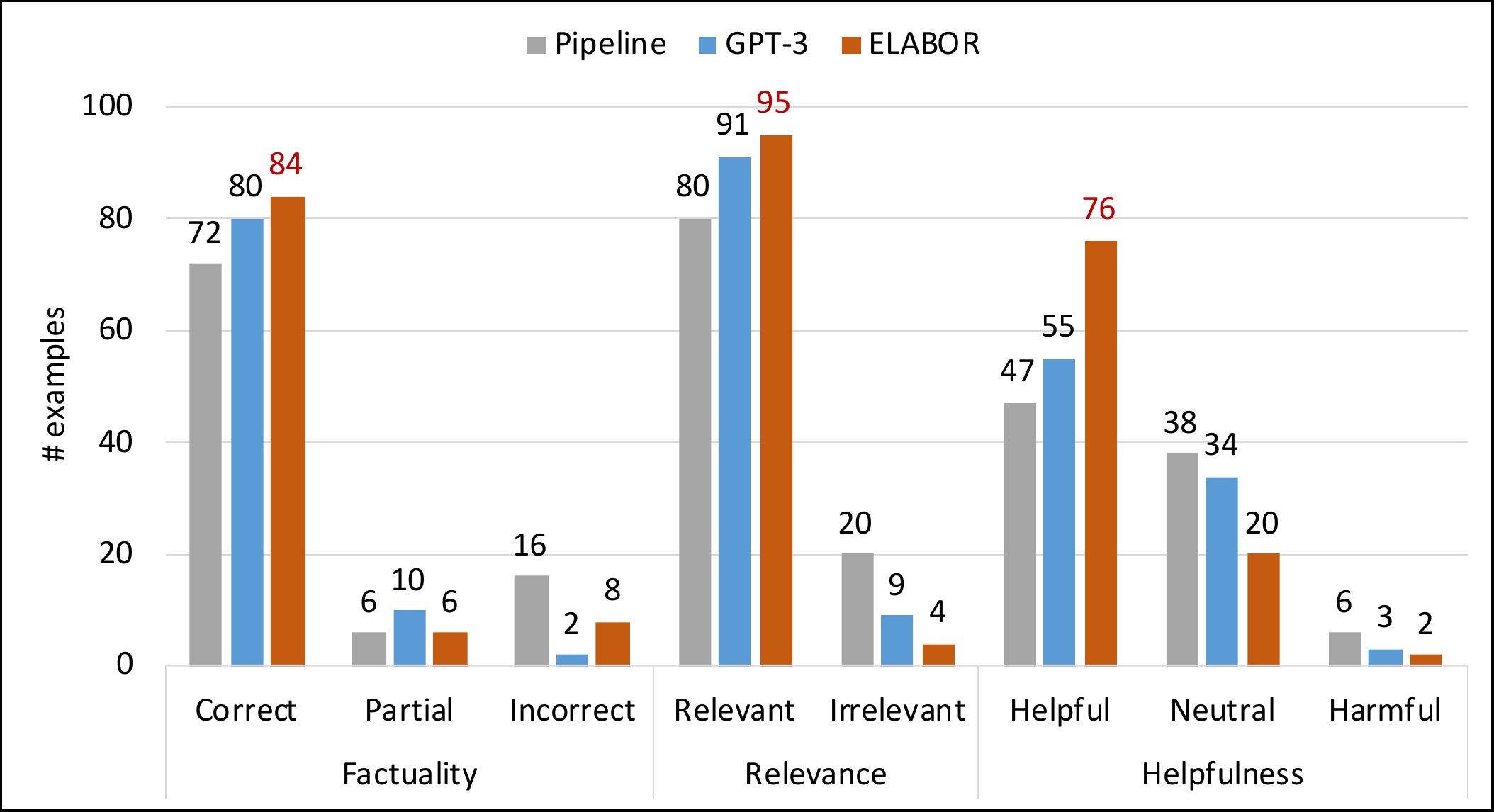}
\caption{Human evaluations on elaborations generated from the generator (Pipeline/\textsc{ElabOr}/GPT-3) which is finally adopted during inference.}\label{fig:human2}
\end{figure}

\vspace{1mm}
\noindent\textbf{Elaboration Quality.} In another experiment, we compare the quality of the elaboration generators from the pipeline setup, GPT-3 and our proposed model \textsc{ElabOr}. We select only one elaboration generated from each model that gives the highest score of the predicted answer during inference, which is actually adopted to produce the final prediction. Adapting from the metrics provided by \citet{shwartz-etal-2020-unsupervised} and \citet{liu-etal-2022-generated}, given a piece of automatically-generated text, we pick three aspects: (1) \emph{Factuality} evaluates whether the text is entirely correct (factual), partially correct (partial) or entirely incorrect (incorrect); (2) \emph{Relevance} evaluates whether the text is relevant or irrelevant to the topics discussed in the question; (3) \emph{Helpfulness} evaluates whether the text provides useful information that helps answer the question (helpful), has no effect (neutral) or is misleading (harmful). 
The human evaluation results on 100 randomly sampled test examples from CSQA2 are shown in Figure \ref{fig:human2}. Clearly, \textsc{ElabOr} achieves better scores across all the three aspects, with the most evident improvement in terms of helpfulness. 
We additionally evaluate how humans benefit from those elaborations generated from our model. The detailed analysis is presented in Appendix \ref{sec:app-human}. Further analysis on how in general the generations from \textsc{ElabOr} and GPT-3 differ is shown in Appendix \ref{sec:app-difference}.

\begin{table}
\centering
	\begin{adjustbox}{max width=1.0\columnwidth}
\begin{tabular}{l|c|c|c|c}
\hline\hline
Data & Count & No Elaboration  & Random Elaboration  & Helpful Elaboration    \\ \hline
QASC   &  70 & 68.57   &  72.86   &    \textbf{85.71}    \\ \hline
CSQA2   & 76 &  55.26  &   61.84  &   \textbf{71.05}    \\  \hline
\end{tabular}
\end{adjustbox}
\vspace{-2mm}
\caption{Performance of \textsc{ElabOr} on 70 and 76 examples picked from 100 human-evaluated instances of QASC dev.~set and CSQA2 dev.~set, respectively, which contain helpful elaborations labeled by workers.}
\label{tbl:human}
\vspace{-4mm}
\end{table}

Based on the annotations given by crowd-sourced workers, we collect only those instances containing an elaboration generated by our model that is labeled as helpful by the workers. This results in 70 and 76 instances from the development set of QASC and CSQA2, respectively. We then compare the performance of \textsc{ElabOr} under three different settings: (1) \emph{No Elaboration} only presents the question to the model during inference; (2) \emph{Random Elaboration} additionally provides a generated elaboration randomly selected after removing the one labeled as helpful; (3) \emph{Helpful Elaboration} contains the single elaboration that is labeled as helpful by workers. The results are shown in Table \ref{tbl:human}. As expected, our model with helpful elaborations outperforms the other two settings by a large margin, aligning with our intuition that meaningful elaborations are beneficial to the task.

\section{Related Work}\label{sec:related}

\noindent\textbf{Direct Inference.} Given only natural-language commonsense questions, a straightforward solution is to directly use language models, either finetuned from the gold-annotated answers \cite{sakaguchi-et-al-winogrande,talmor-etal-2019-commonsenseqa,khashabi-etal-2020-unifiedqa,talmor2021commonsenseqa} or in an unsupervised setting \cite{Trinh2018,petroni-etal-2019-language,puri2019zeroshot,yang-etal-2020-designing,jiang2020} that exploit knowledge already encoded in the pretrained parameters to perform inference. However, beyond the performance score, it is unclear how these models reach the final answer and whether they perform correct reasoning. It is also challenging to conduct direct inference without additional knowledge for complex queries.

\vspace{1mm}
\noindent\textbf{Inference with External Knowledge.} It has been shown that external knowledge such as knowledge bases or Wikipedia contains rich information that could assist inference. Knowledge bases, e.g., ConceptNet \cite{speer-et-al-conceptnet} or ATOMIC \cite{sap-et-al-atomic}, contain relational knowledge that could be incorporated as additional inputs for commonsense QA \cite{Mitra2019,chang-etal-2020-incorporating,BianH0021,ma2021,LvGXTDGSJCH20,yasunaga-etal-2021-qa}. Large corpora are another knowledge source to retrieve question-related facts \cite{lin-etal-2017-reasoning,tandon-etal-2018-reasoning,banerjee-etal-2019-careful,Joshi2020,xiong-etal-2019-improving,Lewis2020}. These knowledge-based approaches depend on the availability and coverage of the knowledge source, which usually depends on the problem domain.

\vspace{1mm}
\noindent\textbf{Inference with Generation.} To alleviate the dependence on external knowledge, recent trends advocate for automatic generation of additional knowledge related to the question via language models. One direction is to learn a generator to generate meaningful justifications for question answering via human-authored explanations \cite{Camburu2018,rajani-etal-2019-explain,Latcinnik2020}. \citet{Bosselut2021DynamicNK} adopted a pretrained commonsense generation model \cite{bosselut-etal-2019-comet} to generate implications of the questions. These approaches, however, require gold-annotated commonsense facts to train a good generator. Another direction explores zero-shot generations using pretrained language models. \citet{shwartz-etal-2020-unsupervised} introduced \emph{Selftalk}, which elicits question clarifications using a few pre-defined templates. \citet{paranjape-etal-2021-prompting} proposed contrastive prompts that compare candidate options for choosing the correct answer. \citet{liu-etal-2022-generated} generated additional texts as continuations of each question by feeding demonstrative prompts to GPT-3. Another work \cite{liu-etal-rainier} used reinforcement learning to guide meaningful generations. \citet{huang-etal-2022-clues} recently proposed to generate clues, which are short phrases or single tokens similar to the gold answers, before answering the question.
Different from existing approaches, we seek to learn an effective generation model jointly with the answer prediction to allow for mutual enhancement.

\section{Conclusion}
We propose a framework for commonsense QA problems that alternates between learning a meaningful, relatively lightweight elaboration generator  and producing an answer from the question and automatically generated elaboration. These two steps are trained interactively, propagating signals to each other. We narrow the performance gap between small LMs and GPT-3, with the elaboration generator producing elaborations judged useful by humans, and matching the performance of the much more expensive GPT-3 model as an elaboration generator. One limitation of \textsc{ElabOr} is lack of exploration beyond GPT-3. We consider investigating this problem as our future work.

\section*{Limitations}
Given the ability of \textsc{ElabOr} to generate free-text elaborations for commonsense question answering, we still observe some cases where the model-generated elaborations are not factually correct, or irrelevant to the question, distracting the answer predictor towards incorrect answers. This reflects a limitation of \textsc{ElabOr} on the controllability of its generations, which is also commonly discovered when using language models for text generation. We consider this as a possible future direction which aims at verifying the factuality and relevancy of model-generated texts before incorporating them for final inference or as a controlling mechanism during generation.

\section*{Ethics \& Broader Impact}
In this work, we only experiment with publicly available datasets. For human evaluation, we do not have access to or collect any personal information from our crowd-sourced workers, except that we only restrict participants to be located in English-speaking countries and have higher qualifications in terms of approval rate. As we work on language model generations, it is possible that the model could produce unintended toxic contents that impede its safe deployment \cite{gehman-etal-2020-realtoxicityprompts}. We do not address this issue here but leave it to the field of controlled generation and language detoxicity.

\section*{Acknowledgments}
The authors appreciate helpful feedback from the anonymous reviewers. We thank Jiacheng Liu for helpful discussions, and the members of H2lab and ARK lab for their constructive feedback. This work was funded in part by the DARPA MCS program through NIWC Pacific (N66001-19-2- 4031), NSF IIS-2044660 and NSF III-2007398. It was also supported by International Postdoctoral Fellowship, Nanyang Technological University.

\bibliography{anthology,custom}

\begin{thebibliography}{53}
\expandafter\ifx\csname natexlab\endcsname\relax\def\natexlab#1{#1}\fi

\bibitem[{Banerjee et~al.(2019)Banerjee, Pal, Mitra, and
  Baral}]{banerjee-etal-2019-careful}
Pratyay Banerjee, Kuntal~Kumar Pal, Arindam Mitra, and Chitta Baral. 2019.
\newblock \href {https://aclanthology.org/P19-1615} {Careful selection of
  knowledge to solve open book question answering}.
\newblock In \emph{Proceedings of the 57th Annual Meeting of the Association
  for Computational Linguistics}, pages 6120--6129. Association for
  Computational Linguistics.

\bibitem[{Bian et~al.(2021)Bian, Han, Chen, and Sun}]{BianH0021}
Ning Bian, Xianpei Han, Bo~Chen, and Le~Sun. 2021.
\newblock \href {https://ojs.aaai.org/index.php/AAAI/article/view/17490}
  {Benchmarking knowledge-enhanced commonsense question answering via
  knowledge-to-text transformation}.
\newblock In \emph{Thirty-Fifth {AAAI} Conference on Artificial Intelligence},
  pages 12574--12582. {AAAI} Press.

\bibitem[{Bosselut et~al.(2021)Bosselut, Bras, and
  Choi}]{Bosselut2021DynamicNK}
Antoine Bosselut, Ronan~Le Bras, and Yejin Choi. 2021.
\newblock Dynamic neuro-symbolic knowledge graph construction for zero-shot
  commonsense question answering.
\newblock In \emph{AAAI}.

\bibitem[{Bosselut et~al.(2019)Bosselut, Rashkin, Sap, Malaviya, Celikyilmaz,
  and Choi}]{bosselut-etal-2019-comet}
Antoine Bosselut, Hannah Rashkin, Maarten Sap, Chaitanya Malaviya, Asli
  Celikyilmaz, and Yejin Choi. 2019.
\newblock \href {https://aclanthology.org/P19-1470} {{COMET}: Commonsense
  transformers for automatic knowledge graph construction}.
\newblock In \emph{Proceedings of the 57th Annual Meeting of the Association
  for Computational Linguistics}, pages 4762--4779.

\bibitem[{Brown et~al.(2020)Brown, Mann, Ryder, Subbiah, Kaplan, Dhariwal,
  Neelakantan, Shyam, Sastry, Askell, Agarwal, Herbert-Voss, Krueger, Henighan,
  Child, Ramesh, Ziegler, Wu, Winter, Hesse, Chen, Sigler, Litwin, Gray, Chess,
  Clark, Berner, McCandlish, Radford, Sutskever, and Amodei}]{tom-et-al-gpt3}
Tom Brown, Benjamin Mann, Nick Ryder, Melanie Subbiah, Jared~D Kaplan, Prafulla
  Dhariwal, Arvind Neelakantan, Pranav Shyam, Girish Sastry, Amanda Askell,
  Sandhini Agarwal, Ariel Herbert-Voss, Gretchen Krueger, Tom Henighan, Rewon
  Child, Aditya Ramesh, Daniel Ziegler, Jeffrey Wu, Clemens Winter, Chris
  Hesse, Mark Chen, Eric Sigler, Mateusz Litwin, Scott Gray, Benjamin Chess,
  Jack Clark, Christopher Berner, Sam McCandlish, Alec Radford, Ilya Sutskever,
  and Dario Amodei. 2020.
\newblock \href
  {https://proceedings.neurips.cc/paper/2020/file/1457c0d6bfcb4967418bfb8ac142f64a-Paper.pdf}
  {Language models are few-shot learners}.
\newblock In \emph{Advances in Neural Information Processing Systems},
  volume~33, pages 1877--1901.

\bibitem[{Camburu et~al.(2018)Camburu, Rockt\"{a}schel, Lukasiewicz, and
  Blunsom}]{Camburu2018}
Oana-Maria Camburu, Tim Rockt\"{a}schel, Thomas Lukasiewicz, and Phil Blunsom.
  2018.
\newblock \href
  {https://proceedings.neurips.cc/paper/2018/file/4c7a167bb329bd92580a99ce422d6fa6-Paper.pdf}
  {e-snli: Natural language inference with natural language explanations}.
\newblock In \emph{Advances in Neural Information Processing Systems},
  volume~31.

\bibitem[{Chang et~al.(2020)Chang, Liu, Gopalakrishnan, Hedayatnia, Zhou, and
  Hakkani-Tur}]{chang-etal-2020-incorporating}
Ting-Yun Chang, Yang Liu, Karthik Gopalakrishnan, Behnam Hedayatnia, Pei Zhou,
  and Dilek Hakkani-Tur. 2020.
\newblock \href {https://aclanthology.org/2020.deelio-1.9} {Incorporating
  commonsense knowledge graph in pretrained models for social commonsense
  tasks}.
\newblock In \emph{Proceedings of Deep Learning Inside Out (DeeLIO): The First
  Workshop on Knowledge Extraction and Integration for Deep Learning
  Architectures}, pages 74--79. Association for Computational Linguistics.

\bibitem[{Chen et~al.(2017)Chen, Fisch, Weston, and
  Bordes}]{chen-etal-2017-reading}
Danqi Chen, Adam Fisch, Jason Weston, and Antoine Bordes. 2017.
\newblock \href {https://aclanthology.org/P17-1171} {Reading {W}ikipedia to
  answer open-domain questions}.
\newblock In \emph{Proceedings of the 55th Annual Meeting of the Association
  for Computational Linguistics (Volume 1: Long Papers)}, pages 1870--1879.

\bibitem[{Devlin et~al.(2019)Devlin, Chang, Lee, and
  Toutanova}]{devlin-etal-2019-bert}
Jacob Devlin, Ming-Wei Chang, Kenton Lee, and Kristina Toutanova. 2019.
\newblock \href {https://aclanthology.org/N19-1423} {{BERT}: Pre-training of
  deep bidirectional transformers for language understanding}.
\newblock In \emph{Proceedings of the 2019 Conference of the North {A}merican
  Chapter of the Association for Computational Linguistics: Human Language
  Technologies, Volume 1 (Long and Short Papers)}, pages 4171--4186.
  Association for Computational Linguistics.

\bibitem[{Gehman et~al.(2020)Gehman, Gururangan, Sap, Choi, and
  Smith}]{gehman-etal-2020-realtoxicityprompts}
Samuel Gehman, Suchin Gururangan, Maarten Sap, Yejin Choi, and Noah~A. Smith.
  2020.
\newblock {R}eal{T}oxicity{P}rompts: Evaluating neural toxic degeneration in
  language models.
\newblock In \emph{Findings of the Association for Computational Linguistics:
  EMNLP 2020}, pages 3356--3369.

\bibitem[{Hinton et~al.(2015)Hinton, Vinyals, and Dean}]{hinton2015distilling}
Geoffrey Hinton, Oriol Vinyals, and Jeff Dean. 2015.
\newblock \href {http://arxiv.org/abs/1503.02531} {Distilling the knowledge in
  a neural network}.

\bibitem[{Holtzman et~al.(2020)Holtzman, Buys, Du, Forbes, and
  Choi}]{Holtzman2020The}
Ari Holtzman, Jan Buys, Li~Du, Maxwell Forbes, and Yejin Choi. 2020.
\newblock \href {https://openreview.net/forum?id=rygGQyrFvH} {The curious case
  of neural text degeneration}.
\newblock In \emph{International Conference on Learning Representations}.

\bibitem[{Huang et~al.(2022)Huang, Wu, Zhou, Gu, Zhao, and
  Cheng}]{huang-etal-2022-clues}
Zixian Huang, Ao~Wu, Jiaying Zhou, Yu~Gu, Yue Zhao, and Gong Cheng. 2022.
\newblock Clues before answers: Generation-enhanced multiple-choice {QA}.
\newblock In \emph{Proceedings of the 2022 Conference of the North American
  Chapter of the Association for Computational Linguistics: Human Language
  Technologies}, pages 3272--3287.

\bibitem[{Jiang et~al.(2020)Jiang, Xu, Araki, and Neubig}]{jiang2020}
Zhengbao Jiang, Frank~F. Xu, Jun Araki, and Graham Neubig. 2020.
\newblock \href {https://doi.org/10.1162/tacl\_a\_00324} {{How Can We Know What
  Language Models Know?}}
\newblock \emph{Transactions of the Association for Computational Linguistics},
  8:423--438.

\bibitem[{Joshi et~al.(2020)Joshi, Lee, Luan, and Toutanova}]{Joshi2020}
Mandar Joshi, Kenton Lee, Yi~Luan, and Kristina Toutanova. 2020.
\newblock \href {http://arxiv.org/abs/2004.12006} {Contextualized
  representations using textual encyclopedic knowledge}.
\newblock \emph{CoRR}, abs/2004.12006.

\bibitem[{Kang et~al.(2020)Kang, Hong, Puerto San~Roman, and
  Myaeng}]{kang-etal-2020-regularization}
Junmo Kang, Giwon Hong, Haritz Puerto San~Roman, and Sung-Hyon Myaeng. 2020.
\newblock Regularization of distinct strategies for unsupervised question
  generation.
\newblock In \emph{Findings of the Association for Computational Linguistics:
  EMNLP 2020}, pages 3266--3277.

\bibitem[{Khashabi et~al.(2020)Khashabi, Min, Khot, Sabharwal, Tafjord, Clark,
  and Hajishirzi}]{khashabi-etal-2020-unifiedqa}
Daniel Khashabi, Sewon Min, Tushar Khot, Ashish Sabharwal, Oyvind Tafjord,
  Peter Clark, and Hannaneh Hajishirzi. 2020.
\newblock \href {https://aclanthology.org/2020.findings-emnlp.171}
  {{UNIFIEDQA}: Crossing format boundaries with a single {QA} system}.
\newblock In \emph{Findings of the Association for Computational Linguistics:
  EMNLP 2020}, pages 1896--1907. Association for Computational Linguistics.

\bibitem[{Khot et~al.(2020)Khot, Clark, Guerquin, Jansen, and
  Sabharwal}]{knot-et-al-qasc}
Tushar Khot, Peter Clark, Michal Guerquin, Peter Jansen, and Ashish Sabharwal.
  2020.
\newblock \href {https://ojs.aaai.org/index.php/AAAI/article/view/6319}
  {{QASC:} {A} dataset for question answering via sentence composition}.
\newblock In \emph{The Thirty-Fourth {AAAI} Conference on Artificial
  Intelligence}, pages 8082--8090. {AAAI} Press.

\bibitem[{Latcinnik and Berant(2020)}]{Latcinnik2020}
Veronica Latcinnik and Jonathan Berant. 2020.
\newblock \href {http://arxiv.org/abs/2004.05569} {Explaining question
  answering models through text generation}.
\newblock \emph{CoRR}, abs/2004.05569.

\bibitem[{Lewis et~al.(2020{\natexlab{a}})Lewis, Liu, Goyal, Ghazvininejad,
  Mohamed, Levy, Stoyanov, and Zettlemoyer}]{lewis-etal-2020-bart}
Mike Lewis, Yinhan Liu, Naman Goyal, Marjan Ghazvininejad, Abdelrahman Mohamed,
  Omer Levy, Veselin Stoyanov, and Luke Zettlemoyer. 2020{\natexlab{a}}.
\newblock \href {https://aclanthology.org/2020.acl-main.703} {{BART}: Denoising
  sequence-to-sequence pre-training for natural language generation,
  translation, and comprehension}.
\newblock In \emph{Proceedings of the 58th Annual Meeting of the Association
  for Computational Linguistics}, pages 7871--7880. Association for
  Computational Linguistics.

\bibitem[{Lewis et~al.(2020{\natexlab{b}})Lewis, Perez, Piktus, Petroni,
  Karpukhin, Goyal, K\"{u}ttler, Lewis, Yih, Rockt\"{a}schel, Riedel, and
  Kiela}]{Lewis2020}
Patrick Lewis, Ethan Perez, Aleksandra Piktus, Fabio Petroni, Vladimir
  Karpukhin, Naman Goyal, Heinrich K\"{u}ttler, Mike Lewis, Wen-tau Yih, Tim
  Rockt\"{a}schel, Sebastian Riedel, and Douwe Kiela. 2020{\natexlab{b}}.
\newblock Retrieval-augmented generation for knowledge-intensive nlp tasks.
\newblock In \emph{Advances in Neural Information Processing Systems},
  volume~33, pages 9459--9474.

\bibitem[{Lin et~al.(2019)Lin, Chen, Chen, and Ren}]{lin-etal-2019-kagnet}
Bill~Yuchen Lin, Xinyue Chen, Jamin Chen, and Xiang Ren. 2019.
\newblock \href {https://aclanthology.org/D19-1282} {{K}ag{N}et:
  Knowledge-aware graph networks for commonsense reasoning}.
\newblock In \emph{Proceedings of the 2019 Conference on Empirical Methods in
  Natural Language Processing and the 9th International Joint Conference on
  Natural Language Processing (EMNLP-IJCNLP)}, pages 2829--2839. Association
  for Computational Linguistics.

\bibitem[{Lin et~al.(2017)Lin, Sun, and Han}]{lin-etal-2017-reasoning}
Hongyu Lin, Le~Sun, and Xianpei Han. 2017.
\newblock \href {https://aclanthology.org/D17-1216} {Reasoning with
  heterogeneous knowledge for commonsense machine comprehension}.
\newblock In \emph{Proceedings of the 2017 Conference on Empirical Methods in
  Natural Language Processing}, pages 2032--2043. Association for Computational
  Linguistics.

\bibitem[{Liu et~al.(2022{\natexlab{a}})Liu, Hallinan, Lu, He, Welleck,
  Hajishirzi, and Choi}]{liu-etal-rainier}
Jiacheng Liu, Skyler Hallinan, Ximing Lu, Pengfei He, Sean Welleck, Hannaneh
  Hajishirzi, and Yejin Choi. 2022{\natexlab{a}}.
\newblock Rainier: Reinforced knowledge introspector for commonsense question
  answering.
\newblock In \emph{Proceedings of the 2022 Conference on Empirical Methods in
  Natural Language Processing (EMNLP)}.

\bibitem[{Liu et~al.(2022{\natexlab{b}})Liu, Liu, Lu, Welleck, West, Le~Bras,
  Choi, and Hajishirzi}]{liu-etal-2022-generated}
Jiacheng Liu, Alisa Liu, Ximing Lu, Sean Welleck, Peter West, Ronan Le~Bras,
  Yejin Choi, and Hannaneh Hajishirzi. 2022{\natexlab{b}}.
\newblock \href {https://aclanthology.org/2022.acl-long.225} {Generated
  knowledge prompting for commonsense reasoning}.
\newblock In \emph{Proceedings of the 60th Annual Meeting of the Association
  for Computational Linguistics (Volume 1: Long Papers)}, pages 3154--3169.
  Association for Computational Linguistics.

\bibitem[{Liu et~al.(2019)Liu, Ott, Goyal, Du, Joshi, Chen, Levy, Lewis,
  Zettlemoyer, and Stoyanov}]{liu-et-al-roberta}
Yinhan Liu, Myle Ott, Naman Goyal, Jingfei Du, Mandar Joshi, Danqi Chen, Omer
  Levy, Mike Lewis, Luke Zettlemoyer, and Veselin Stoyanov. 2019.
\newblock \href {http://arxiv.org/abs/1907.11692} {Roberta: {A} robustly
  optimized {BERT} pretraining approach}.
\newblock \emph{CoRR}, abs/1907.11692.

\bibitem[{Lv et~al.(2020)Lv, Guo, Xu, Tang, Duan, Gong, Shou, Jiang, Cao, and
  Hu}]{LvGXTDGSJCH20}
Shangwen Lv, Daya Guo, Jingjing Xu, Duyu Tang, Nan Duan, Ming Gong, Linjun
  Shou, Daxin Jiang, Guihong Cao, and Songlin Hu. 2020.
\newblock Graph-based reasoning over heterogeneous external knowledge for
  commonsense question answering.
\newblock In \emph{{AAAI}}, pages 8449--8456. {AAAI} Press.

\bibitem[{Ma et~al.(2021)Ma, Ilievski, Francis, Bisk, Nyberg, and
  Oltramari}]{ma2021}
Kaixin Ma, Filip Ilievski, Jonathan Francis, Yonatan Bisk, Eric Nyberg, and
  Alessandro Oltramari. 2021.
\newblock Knowledge-driven data construction for zero-shot evaluation in
  commonsense question answering.
\newblock In \emph{{AAAI}}, pages 13507--13515. {AAAI} Press.

\bibitem[{Mihaylov et~al.(2018)Mihaylov, Clark, Khot, and
  Sabharwal}]{mihaylov-etal-2018-suit}
Todor Mihaylov, Peter Clark, Tushar Khot, and Ashish Sabharwal. 2018.
\newblock \href {https://www.aclweb.org/anthology/D18-1260} {Can a suit of
  armor conduct electricity? a new dataset for open book question answering}.
\newblock In \emph{Proceedings of the 2018 Conference on Empirical Methods in
  Natural Language Processing}, pages 2381--2391. Association for Computational
  Linguistics.

\bibitem[{Mihaylov and Frank(2018)}]{mihaylov-frank-2018-knowledgeable}
Todor Mihaylov and Anette Frank. 2018.
\newblock \href {https://aclanthology.org/P18-1076} {Knowledgeable reader:
  Enhancing cloze-style reading comprehension with external commonsense
  knowledge}.
\newblock In \emph{Proceedings of the 56th Annual Meeting of the Association
  for Computational Linguistics (Volume 1: Long Papers)}, pages 821--832.
  Association for Computational Linguistics.

\bibitem[{Min et~al.(2019)Min, Chen, Hajishirzi, and Zettlemoyer}]{MinCHZ19}
Sewon Min, Danqi Chen, Hannaneh Hajishirzi, and Luke Zettlemoyer. 2019.
\newblock \href {https://doi.org/10.18653/v1/D19-1284} {A discrete hard {EM}
  approach for weakly supervised question answering}.
\newblock In \emph{Proceedings of the 2019 Conference on Empirical Methods in
  Natural Language Processing and the 9th International Joint Conference on
  Natural Language Processing, {EMNLP-IJCNLP}}, pages 2851--2864.

\bibitem[{Mitra et~al.(2019)Mitra, Banerjee, Pal, Mishra, and
  Baral}]{Mitra2019}
Arindam Mitra, Pratyay Banerjee, Kuntal~Kumar Pal, Swaroop Mishra, and Chitta
  Baral. 2019.
\newblock \href {http://arxiv.org/abs/1909.08855} {Exploring ways to
  incorporate additional knowledge to improve natural language commonsense
  question answering}.
\newblock \emph{CoRR}, abs/1909.08855.

\bibitem[{Paranjape et~al.(2021)Paranjape, Michael, Ghazvininejad, Hajishirzi,
  and Zettlemoyer}]{paranjape-etal-2021-prompting}
Bhargavi Paranjape, Julian Michael, Marjan Ghazvininejad, Hannaneh Hajishirzi,
  and Luke Zettlemoyer. 2021.
\newblock \href {https://aclanthology.org/2021.findings-acl.366} {Prompting
  contrastive explanations for commonsense reasoning tasks}.
\newblock In \emph{Findings of the Association for Computational Linguistics:
  ACL-IJCNLP 2021}, pages 4179--4192.

\bibitem[{Petroni et~al.(2019)Petroni, Rockt{\"a}schel, Riedel, Lewis, Bakhtin,
  Wu, and Miller}]{petroni-etal-2019-language}
Fabio Petroni, Tim Rockt{\"a}schel, Sebastian Riedel, Patrick Lewis, Anton
  Bakhtin, Yuxiang Wu, and Alexander Miller. 2019.
\newblock \href {https://aclanthology.org/D19-1250} {Language models as
  knowledge bases?}
\newblock In \emph{Proceedings of the 2019 Conference on Empirical Methods in
  Natural Language Processing and the 9th International Joint Conference on
  Natural Language Processing (EMNLP-IJCNLP)}, pages 2463--2473. Association
  for Computational Linguistics.

\bibitem[{Puri and Catanzaro(2019)}]{puri2019zeroshot}
Raul Puri and Bryan Catanzaro. 2019.
\newblock \href {http://arxiv.org/abs/1912.10165} {Zero-shot text
  classification with generative language models}.

\bibitem[{Qu et~al.(2021)Qu, Chen, Xhonneux, Bengio, and Tang}]{qu2021rnnlogic}
Meng Qu, Junkun Chen, Louis-Pascal Xhonneux, Yoshua Bengio, and Jian Tang.
  2021.
\newblock \href {https://openreview.net/forum?id=tGZu6DlbreV}
  {{\{}RNNL{\}}ogic: Learning logic rules for reasoning on knowledge graphs}.
\newblock In \emph{International Conference on Learning Representations}.

\bibitem[{Radford et~al.(2019)Radford, Wu, Child, Luan, Amodei, and
  Sutskever}]{Radford2019}
Alec Radford, Jeffrey Wu, Rewon Child, David Luan, Dario Amodei, and Ilya
  Sutskever. 2019.
\newblock \href {https://openai.com/blog/better-language-models/} {{Language
  Models are Unsupervised Multitask Learners}}.

\bibitem[{Raffel et~al.(2020)Raffel, Shazeer, Roberts, Lee, Narang, Matena,
  Zhou, Li, and Liu}]{raffel-et-al-t5}
Colin Raffel, Noam Shazeer, Adam Roberts, Katherine Lee, Sharan Narang, Michael
  Matena, Yanqi Zhou, Wei Li, and Peter~J. Liu. 2020.
\newblock \href {http://jmlr.org/papers/v21/20-074.html} {Exploring the limits
  of transfer learning with a unified text-to-text transformer}.
\newblock \emph{Journal of Machine Learning Research}, 21(140):1--67.

\bibitem[{Rajani et~al.(2019)Rajani, McCann, Xiong, and
  Socher}]{rajani-etal-2019-explain}
Nazneen~Fatema Rajani, Bryan McCann, Caiming Xiong, and Richard Socher. 2019.
\newblock \href {https://aclanthology.org/P19-1487} {Explain yourself!
  leveraging language models for commonsense reasoning}.
\newblock In \emph{Proceedings of the 57th Annual Meeting of the Association
  for Computational Linguistics}, pages 4932--4942. Association for
  Computational Linguistics.

\bibitem[{Reimers and Gurevych(2019)}]{reimers-gurevych-2019-sentence}
Nils Reimers and Iryna Gurevych. 2019.
\newblock \href {https://aclanthology.org/D19-1410} {Sentence-{BERT}: Sentence
  embeddings using {S}iamese {BERT}-networks}.
\newblock In \emph{Proceedings of the 2019 Conference on Empirical Methods in
  Natural Language Processing and the 9th International Joint Conference on
  Natural Language Processing (EMNLP-IJCNLP)}, pages 3982--3992, Hong Kong,
  China.

\bibitem[{Sakaguchi et~al.(2021)Sakaguchi, Bras, Bhagavatula, and
  Choi}]{sakaguchi-et-al-winogrande}
Keisuke Sakaguchi, Ronan~Le Bras, Chandra Bhagavatula, and Yejin Choi. 2021.
\newblock \href {https://doi.org/10.1145/3474381} {Winogrande: An adversarial
  winograd schema challenge at scale}.
\newblock \emph{Commun. ACM}, 64(9):99–106.

\bibitem[{Sap et~al.(2019)Sap, Bras, Allaway, Bhagavatula, Lourie, Rashkin,
  Roof, Smith, and Choi}]{sap-et-al-atomic}
Maarten Sap, Ronan~Le Bras, Emily Allaway, Chandra Bhagavatula, Nicholas
  Lourie, Hannah Rashkin, Brendan Roof, Noah~A. Smith, and Yejin Choi. 2019.
\newblock \href {https://doi.org/10.1609/aaai.v33i01.33013027} {{ATOMIC:} an
  atlas of machine commonsense for if-then reasoning}.
\newblock In \emph{The Thirty-Third {AAAI} Conference on Artificial
  Intelligence}, pages 3027--3035. {AAAI} Press.

\bibitem[{Shwartz et~al.(2020)Shwartz, West, Le~Bras, Bhagavatula, and
  Choi}]{shwartz-etal-2020-unsupervised}
Vered Shwartz, Peter West, Ronan Le~Bras, Chandra Bhagavatula, and Yejin Choi.
  2020.
\newblock \href {https://aclanthology.org/2020.emnlp-main.373} {Unsupervised
  commonsense question answering with self-talk}.
\newblock In \emph{Proceedings of the 2020 Conference on Empirical Methods in
  Natural Language Processing (EMNLP)}, pages 4615--4629. Association for
  Computational Linguistics.

\bibitem[{Speer et~al.(2017)Speer, Chin, and Havasi}]{speer-et-al-conceptnet}
Robyn Speer, Joshua Chin, and Catherine Havasi. 2017.
\newblock Conceptnet 5.5: An open multilingual graph of general knowledge.
\newblock In \emph{Proceedings of the Thirty-First AAAI Conference on
  Artificial Intelligence}, page 4444–4451. AAAI Press.

\bibitem[{Talmor et~al.(2019)Talmor, Herzig, Lourie, and
  Berant}]{talmor-etal-2019-commonsenseqa}
Alon Talmor, Jonathan Herzig, Nicholas Lourie, and Jonathan Berant. 2019.
\newblock \href {https://aclanthology.org/N19-1421} {{C}ommonsense{QA}: A
  question answering challenge targeting commonsense knowledge}.
\newblock In \emph{Proceedings of the 2019 Conference of the North {A}merican
  Chapter of the Association for Computational Linguistics: Human Language
  Technologies, Volume 1 (Long and Short Papers)}, pages 4149--4158.
  Association for Computational Linguistics.

\bibitem[{Talmor et~al.(2021)Talmor, Yoran, Bras, Bhagavatula, Goldberg, Choi,
  and Berant}]{talmor2021commonsenseqa}
Alon Talmor, Ori Yoran, Ronan~Le Bras, Chandra Bhagavatula, Yoav Goldberg,
  Yejin Choi, and Jonathan Berant. 2021.
\newblock \href {https://openreview.net/forum?id=qF7FlUT5dxa} {Commonsense{QA}
  2.0: Exposing the limits of {AI} through gamification}.
\newblock In \emph{Thirty-fifth Conference on Neural Information Processing
  Systems Datasets and Benchmarks Track (Round 1)}.

\bibitem[{Tandon et~al.(2018)Tandon, Dalvi, Grus, Yih, Bosselut, and
  Clark}]{tandon-etal-2018-reasoning}
Niket Tandon, Bhavana Dalvi, Joel Grus, Wen-tau Yih, Antoine Bosselut, and
  Peter Clark. 2018.
\newblock \href {https://aclanthology.org/D18-1006} {Reasoning about actions
  and state changes by injecting commonsense knowledge}.
\newblock In \emph{Proceedings of the 2018 Conference on Empirical Methods in
  Natural Language Processing}, pages 57--66. Association for Computational
  Linguistics.

\bibitem[{Trinh and Le(2018)}]{Trinh2018}
Trieu~H. Trinh and Quoc~V. Le. 2018.
\newblock \href {http://arxiv.org/abs/1806.02847} {A simple method for
  commonsense reasoning}.
\newblock \emph{CoRR}, abs/1806.02847.

\bibitem[{West et~al.(2021)West, Bhagavatula, Hessel, Hwang, Jiang, Bras, Lu,
  Welleck, and Choi}]{West2021SymbolicKD}
Peter West, Chandrasekhar Bhagavatula, Jack Hessel, Jena~D. Hwang, Liwei Jiang,
  Ronan~Le Bras, Ximing Lu, Sean Welleck, and Yejin Choi. 2021.
\newblock Symbolic knowledge distillation: from general language models to
  commonsense models.
\newblock \emph{ArXiv}, abs/2110.07178.

\bibitem[{Williams(1992)}]{Williams:92}
Ronald~J. Williams. 1992.
\newblock Simple statistical gradient-following algorithms for connectionist
  reinforcement learning.
\newblock \emph{Machine Learning}, 8:229--256.

\bibitem[{Xiong et~al.(2019)Xiong, Yu, Chang, Guo, and
  Wang}]{xiong-etal-2019-improving}
Wenhan Xiong, Mo~Yu, Shiyu Chang, Xiaoxiao Guo, and William~Yang Wang. 2019.
\newblock \href {https://aclanthology.org/P19-1417} {Improving question
  answering over incomplete {KB}s with knowledge-aware reader}.
\newblock In \emph{Proceedings of the 57th Annual Meeting of the Association
  for Computational Linguistics}, pages 4258--4264. Association for
  Computational Linguistics.

\bibitem[{Yang et~al.(2020)Yang, Lin, Nogueira, Tsai, Wang, and
  Lin}]{yang-etal-2020-designing}
Jheng-Hong Yang, Sheng-Chieh Lin, Rodrigo Nogueira, Ming-Feng Tsai, Chuan-Ju
  Wang, and Jimmy Lin. 2020.
\newblock \href {https://aclanthology.org/2020.coling-main.307} {Designing
  templates for eliciting commonsense knowledge from pretrained
  sequence-to-sequence models}.
\newblock In \emph{Proceedings of the 28th International Conference on
  Computational Linguistics}, pages 3449--3453.

\bibitem[{Yasunaga et~al.(2021)Yasunaga, Ren, Bosselut, Liang, and
  Leskovec}]{yasunaga-etal-2021-qa}
Michihiro Yasunaga, Hongyu Ren, Antoine Bosselut, Percy Liang, and Jure
  Leskovec. 2021.
\newblock \href {https://aclanthology.org/2021.naacl-main.45} {{QA}-{GNN}:
  Reasoning with language models and knowledge graphs for question answering}.
\newblock In \emph{Proceedings of the 2021 Conference of the North American
  Chapter of the Association for Computational Linguistics: Human Language
  Technologies}, pages 535--546. Association for Computational Linguistics.

\end{thebibliography}
\bibliographystyle{acl_natbib}

\appendix

\section{Appendix}\label{sec:appendix}


\subsection{Algorithm}\label{sec:app-algo}
The overall algorithm for training \textsc{ElabOr} is shown in Algorithm~\ref{alg}.

\begin{algorithm}[h!]\label{alg}
\small
	\caption{Training procedure of \textsc{ElabOr}.}\label{alg}
	\begin{algorithmic}[1]
		\STATE {\bf Initialize:} For each question $q$, use GPT-3 to sample a set of knowledge $\bar{\mathcal{E}}$ as continuations of $q$ (Section \ref{sec:distill}).
		\FOR {epoch$=1,...,T$}
		    \FOR {batch$=1,...,N$}
		    \STATE Optimize Eq.~\ref{eqn:expectation} by alternating between {\bf A} and {\bf B}:
		    \vspace{1mm}
		    \STATE {\bf A.} Optimize elaboration generator $\mathcal{F}_E$ to produce $P(e|q)$ (Section \ref{sec:em})
		        \FOR {a question-answer pair $(q,a)$ in batch}
		        \vspace{1mm}
		        \STATE {\bf E-Step:} Select top-$K$ elaborations $\mathcal{E}=\{e_1,...,e_K\} \subseteq \bar{\mathcal{E}}$ given scores produced from the answer predictor.
		        \STATE {\bf M-Step:} Update the elaboration generator $\mathcal{F}_E$ using Eq.~\ref{eqn:maxFc} with $\mathcal{E}$ and $q$.
		        \ENDFOR
		    
		    \vspace{1mm}   
		    \STATE {\bf B.} Optimize answer predictor $\mathcal{F}_A$ to produce $P(a\mid e,q)$ (Section \ref{sec:answer}) 
		        \FOR {a question-answer pair $(q,a)$ in batch}
		        \STATE Sample a set of candidate elaborations $\tilde{\mathcal{E}}$ using $\mathcal{F}_E$ trained in the previous step.
		        \STATE For each $\tilde{e}\in\tilde{\mathcal{E}}$, update the answer predictor $\mathcal{F}_A$ by maximizing Eq.~\ref{eqn:predictor} given $a$ and $\tilde{e}$.
		        \ENDFOR
		
		    \ENDFOR
		\ENDFOR
	\end{algorithmic}
\end{algorithm}

\begin{table*}
\footnotesize
\centering
	\begin{adjustbox}{max width=1.0\textwidth}
\begin{tabular}{l|l|l}
\hline
\textbf{Question} & \textbf{Elaboration} & \textbf{Answer} \\ \hline
What does your ear drum do when it hears & The ear drum is the part of the human body that is responsible & \multirow{2}{*}{Vibrates} \\ 
something? & for hearing. When you hear something, the ear drum vibrates. & \\
\hline 
How can we find out how much something & Weighing is done by using a scale. The amount of matter in & \multirow{2}{*}{using a scale} \\
weighs? & an object is measured by weighing it. & \\
\hline
The period of most rapid growth after birth  &  \multirow{2}{*}{The period of fastest growth is in the first few weeks.} & \multirow{2}{*}{a baby} \\ is when they are what? & & \\
\hline
\multirow{2}{*}{What does predicting weather require?} & Weathering prediction requires observation of weather conditions. & \multirow{2}{*}{meterologists} \\
& Forecasting weather requires observing weather patterns and clouds. & \\
\hline
A polar bear does what to survive in its & Polar bears have thick fur to keep them warm. They are able to & \multirow{2}{*}{grows fur} \\
environment? & swim and hunt for food. Polar bears live in cold areas. & \\
\hline
Seismographs measure what aspect of & Seismographs measure the height and direction of earthquakes. & \multirow{2}{*}{magnitude} \\
earthquakes? & The seismic wave is measured by seismographs. & \\
\hline
\multirow{2}{*}{What decreases tooth decay?} & The use of fluoride in drinking water is used to decrease tooth & \multirow{2}{*}{drinking water} \\
& decay. Fluoride is added to the water to prevent it from decaying. & \\
\hline
Some pelycosaurs gave rise to reptile & Amphibians and mammals are both examples of animals that have & \multirow{2}{*}{mammals} \\
ancestral to? & reptilian characteristics. & \\
\hline
\multirow{2}{*}{Your polygenic traits determine?} & Polygenic traits are inherited. The trait that determines your color & if you are \\
& is your genes. & white or brown \\
\hline
\end{tabular}
\end{adjustbox}
\caption{Generated elaborations from our learned generator GPT2-large}\label{tbl:generations}
\end{table*}

\subsection{Data \& Experimental Setup}\label{sec:setup}
(1) \textbf{CommonsenseQA} (CSQA; \citealp{talmor-etal-2019-commonsenseqa}) is created based on commonsense knowledge from various concepts in ConceptNet. Most of the questions require implicit background knowledge that is trivial to humans. The dataset consists of 12,247 examples (80\%/10\%/10\% train/dev./test split), each of which is a 5-way multiple-choice selection problem. (2) \textbf{CommonsenseQA 2.0} (CSQA2; \citealp{talmor2021commonsenseqa}) is a more challenging dataset collected in an adversarial manner where a user is encouraged to create questions for which a well-trained \textsc{ROBERTA} model \cite{liu-et-al-roberta} fails to provide the correct answer. The dataset contains a total of 14,343 questions (9,282 train, 2,544 dev., 2,517 test) with binary answer choices (yes/no). (3) \textbf{QASC} \cite{knot-et-al-qasc} is a question answering dataset requiring compositions of multiple pieces of texts. It is collected from elementary and middle-school science questions. The dataset contains 9,980 questions (8,134 train, 926 dev., 920 test), each of which is followed by 8 different choices. Note that we do not use the gold-annotated background facts accompanied with the original data, in order to test the model's ability to automatically elicit knowledge and reason. (4) \textbf{OpenBookQA} (OBQA; \citealp{mihaylov-etal-2018-suit}) is a collection of open book exams on elementary-level science facts. It contains a total of 5,957 questions (4,957 train, 500 dev., 500 test) with four candidate choices for each question. Similar to QASC, we also remove the gold-annotated science facts in the original release.

For experimental setup, we use GPT-3 \cite{tom-et-al-gpt3} under few-shot prompting and with nucleus sampling
$p=0.5$ \cite{Holtzman2020The} to sample 20 elaborations for each question. We use the same prompts as those from \citet{liu-etal-2022-generated} and provide them in Table~\ref{tbl:prompt}. During alternative training, for each iteration, we use 100 instances to update the elaboration generator followed by the answer predictor. We adopt Adam optimizer with learning rate initialized at $10^{-5}$ 
for both components. The elaboration generator generates $|\tilde{\mathcal{E}}|=10$ elaborations during both training and testing phases via nucleus sampling
$p=0.95$ and with temperature set as $0.7$. We set $K=3$ when forming the  top-$K$ elaboration set $\bar{\mathcal{E}}$ during the  E-step. For elaboration generation, GPT2-large and BART-large has 774M and 406M parameters, respectively. For answer prediction, we use T5 with varying model sizes: 770M for T5-large/UnifiedQA-large and 3B for UnifiedQA-3b. 

\subsection{Generations from \textsc{ElabOr}}\label{sec:app-generations}
We list some actual generations from \textsc{ElabOr} using the learned elaboration generator GPT2-large in Table~\ref{tbl:generations}. These examples are selected from those used for human evaluations. The listed elaboration for each question is the most confident elaboration that is used for final prediction.

\begin{table*}
\footnotesize
\centering
	\begin{adjustbox}{max width=1.0\textwidth}
\begin{tabular}{l|l}
\hline
\textbf{Task}  & \textbf{Prompt}  \\ \hline
\multirow{12}{*}{CSQA}   & Generate some knowledge about the concepts in the input. Examples: \\ 
& Input: Google Maps and other highway and street GPS services have replaced what?  \\ 
& \textit{Knowledge: Electronic maps are the modern version of paper atlas.}   \\  
& Input: The fox walked from the city into the forest, what was it looking for?  \\ 
& \textit{Knowledge: Natural habitats are usually away from cities.}   \\
& Input: You can share files with someone if you have a connection to a what? \\ 
& \textit{Knowledge: Files can be shared over the Internet.}   \\
& Input: Too many people want exotic snakes. The demand is driving what to carry them? \\ 
& \textit{Knowledge: Some people raise snakes as pets.}   \\
& Input: The body guard was good at his duties, he made the person who hired him what? \\ 
& \textit{Knowledge: The job of body guards is to ensure the safety and security of the employer}   \\
& Input: \{question\} \\
& Knowledge: \\
\hline
\multirow{20}{*}{CSQA2}   & Generate some knowledge about the input. Examples: \\ 
& Input: Greece is larger than mexico.  \\ 
& \textit{Knowledge: Greece is approximately 131,957 sq km, while Mexico is approximately 1,964,375 sq km, making Mexico}   \\ 
& \textit{1,389\% larger than Greece.} \\
& Input: Glasses always fog up.  \\ 
& \textit{Knowledge: Condensation occurs on eyeglass lenses when water vapor from your sweat, breath, and ambient humidity} \\
& \textit{lands on a cold surface, cools, and then changes into tiny drops of liquid, forming a film that you see as fog. Your lenses} \\
& \textit{will be relatively cool compared to your breath, especially when the outside air is cold.}   \\
& Input: A fish is capable of thinking. \\ 
& \textit{Knowledge: Fish are more intelligent than they appear. In many areas, such as memory, their
cognitive powers match or} \\ 
& \textit{exceed those of `higher' vertebrates including non-human primates. Fish's long-term memories help them keep track of} \\
& \textit{complex social relationships.} \\
& Input: A common effect of smoking lots of cigarettes in one’s lifetime is a higher than
normal chance of getting lung cancer. \\ 
& \textit{Knowledge: Those who consistently averaged less than one cigarette per day over their lifetime had nine times the risk of} \\
& \textit{dying from lung cancer than never smokers. Among people who smoked between one and 10 cigarettes per day, the risk of} \\ 
& \textit{dying from lung cancer was nearly 12 times higher than that of never smokers.}  \\
& Input: A rock is the same size as a pebble. \\ 
& \textit{Knowledge: A pebble is a clast of rock with a particle size of 4 to 64 millimetres based on the Udden-Wentworth scale of} \\
& \textit{sedimentology. Pebbles are generally considered larger than granules (2 to 4 millimetres diameter) and smaller than cobbles}   \\
& \textit{(64 to 256 millimetres diameter).} \\
& Input: \{question\} \\
& Knowledge: \\
\hline
\multirow{12}{*}{QASC}   & Generate some knowledge about the input. Examples: \\ 
& Input: What type of water formation is formed by clouds?  \\ 
& \textit{Knowledge: Clouds are made of water vapor.}   \\  
& Input: What can prevent food spoilage?  \\ 
& \textit{Knowledge: Dehydrating food is used for preserving food}   \\
& Input: The process by which genes are passed is \\
& \textit{Knowledge: Genes are passed from parent to offspring.}   \\
& Input: The stomach does what in the body? \\ 
& \textit{Knowledge: The stomach is part of the digestive system}   \\
& Input: What can cause rocks to break down? \\ 
& \textit{Knowledge: Mechanical weathering is when rocks are broken down by mechanical means.}   \\
& Input: \{question\} \\
& Knowledge: \\
\hline
\multirow{12}{*}{OBQA}   & Generate some knowledge given the question. Examples: \\ 
& Question: Which would likely transfer special heat via waves?  \\ 
& \textit{Knowledge: Radiation is when heat is transferred through waves. Radiation is made by certain bombs.}   \\  
& Question: When standing miles away from Mount Rushmore  \\ 
& \textit{Knowledge: As distance to an object increases, that object will appear smaller.}   \\
& Question: Ducks might their webbed appendages to \\
& \textit{Knowledge: Webbed feet are used for moving faster through water by aquatic animals.}   \\
& Question: Which would a strawberry most rely on to ensure it gets planted? \\ 
& \textit{Knowledge: Birds are a vehicle for spreading the seeds of a plant.}   \\
& Question: A typhoon can potentially cause \\ 
& \textit{Knowledge: A typhoon can bring a lot of rainfall. Heavy rains cause flooding.}   \\
& Input: \{question\} \\
& Knowledge: \\
\hline
\end{tabular}
\end{adjustbox}
\caption{Exact prompts used for each dataset. \{question\} indicates a placeholder for each input question.}\label{tbl:prompt}
\end{table*}

\subsection{Human Evaluation}\label{sec:app-human}
We additionally evaluate how humans benefit from those elaborations generated from our model across 100 random-sampled development examples from QASC. For each example, we first present the workers with the question and ask them to choose only one answer from multiple choices. In another round, we provide both the question and the generated elaboration to the workers and collect their answers. The two rounds of experiments recruit non-overlapping annotators to ensure validity. As a result, 78 questions are correctly answered by workers without seeing extra elaborations. On the other hand, 81 questions are correctly answered when elaborations are provided. This shows our elaboration generator is still beneficial to humans even though commonsense QA appears to be much easier for humans than machines.

\subsection{\textsc{ElabOr} vs. GPT-3}\label{sec:app-difference}
We select 50 examples from those used for human evaluation, half of which are correctly predicted by \textsc{ElabOr} but wrongly predicted by GPT-3 (denoted as D1). In the remaining 25 cases, the situation is the opposite (denoted as D2). Through manual inspection, we observe that in D1, \textsc{ElabOr} is often better off when the question is more general, e.g., ``\textit{What is a simple mode of transportation?}''. \textsc{ElabOr} can generate more specific information relevant to some answer choices and tends to speak more. For D2, \textsc{ElabOr} performs worse when the model overgenerates noisy information not related to the question context leading to wrong answers. For example, the question ``\textit{What do choanocytes have to trap the particles?}'' causes \textsc{ElabOr} to generate ``\textit{The particle is a virus. The choanocytes are part of the immune system. The antibodies that bind the virus and destroy it.}'' which does not answer the question.

\end{document}